\documentclass[runningheads]{llncs}
\usepackage{graphicx}

\usepackage{tikz}
\usepackage{comment}
\usepackage{amsmath,amssymb} %
\usepackage{color}

\usepackage[accsupp]{axessibility}  %

\usepackage{amsmath}

\renewcommand{\vec}[1]{\boldsymbol{#1}}
\newcommand{\mat}[1]{\mathbf{#1}}
\newcommand{\set}[1]{\mathcal{#1}}
\newcommand{\vect}[1]{\mathbf{#1}}

\usepackage[capitalize]{cleveref}
\crefname{section}{Sec.}{Secs.}
\Crefname{section}{Section}{Sections}
\Crefname{table}{Table}{Tables}
\crefname{table}{Tab.}{Tabs.}

\newcommand{\pose}[0]{\vec{\theta}}
\newcommand{\shape}[0]{\vec{\beta}}

\newcommand{\smpl}[0]{H}

\newcommand{\behave}[1]{BEHAVE}

\newcommand{\Rone}[1]{\textcolor{blue}{\textbf{R1}}}
\newcommand{\Rtwo}[1]{\textcolor{orange}{\textbf{R2}}}
\newcommand{\Rthree}[1]{\textcolor{red}{\textbf{R3}}}

\usepackage{epsfig}
\usepackage{graphicx}
\usepackage{comment}
\usepackage{amsmath,amssymb} %
\usepackage{color}
\usepackage[export]{adjustbox}
\usepackage{cite} %
\usepackage{floatrow}
\newfloatcommand{capbtabbox}{table}[][\FBwidth]

\usepackage{wrapfig} %

\def\etal{{\em et al. }}
\usepackage[font=small,skip=5pt]{caption}
\usepackage{float}

\usepackage{multirow}
\usepackage{tabularx}
\usepackage{soul}
\usepackage{array}
\usepackage{comment}

\usepackage[mathscr]{euscript}
\newcolumntype{L}[1]{>{\raggedright\let\newline\\arraybackslash\hspace{0pt}}m{#1}}
\newcolumntype{C}[1]{>{\centering\let\newline\\arraybackslash\hspace{0pt}}m{#1}}
\newcolumntype{R}[1]{>{\raggedleft\let\newline\\arraybackslash\hspace{0pt}}m{#1}}
\newcolumntype{Y}{>{\centering\arraybackslash}X}
\usepackage{booktabs}
\usepackage{enumitem}

\begin{document}
\pagestyle{headings}
\mainmatter
\def\ECCVSubNumber{4894}  %

\title{CHORE: Contact, Human and Object REconstruction from a single RGB image} %

\titlerunning{CHORE}
\author{
\index{Xie, Xianghui}
Xianghui Xie \inst{1}
\and 
\index{Lal Bhatnagar, Bharat}
Bharat Lal Bhatnagar \inst{1,2}
\and
\index{Pons-Moll, Gerard}
Gerard Pons-Moll \inst{1,2}} 
\institute{
Max Planck Institute for Informatics, Saarland Informatics Campus, Germany  \and
University of T\"ubingen, Germany \\ \email{\{xxie, bbhatnag\}@mpi-inf.mpg.de, gerard.pons-moll@uni-tuebingen.de} 
}

\maketitle

\begin{abstract}
While most works in computer vision and learning have
focused on perceiving 3D humans from single images in isolation, in this work we focus on capturing 3D humans interacting with objects. 
The problem is extremely challenging due to heavy occlusions between human and object, diverse interaction types and depth ambiguity. In this paper, we introduce CHORE, a novel method
that learns to jointly reconstruct human and object from
a single image. CHORE takes inspiration from recent advances in implicit surface learning and classical model-based fitting. We
compute a neural reconstruction of human and object represented implicitly with two unsigned distance fields, and
additionally predict a correspondence field to a parametric body as well as an object pose field. This allows us to
robustly fit a parametric body model and a 3D object template, while reasoning about interactions. Furthermore, prior pixel-aligned implicit learning methods use synthetic data and make assumptions that are not met in real data. We propose a simple yet effective depth-aware scaling that allows more efficient shape learning on real data. Our experiments show
that our joint reconstruction learned with the proposed strategy significantly outperforms the SOTA. Our code and models are available at
\textcolor{blue}{\fontfamily{lmss}\selectfont https://virtualhumans.mpi-inf.mpg.de/chore}

\end{abstract}

\section{Introduction}
\begin{figure}[t]
    \centering
    \includegraphics[width=0.95\textwidth]{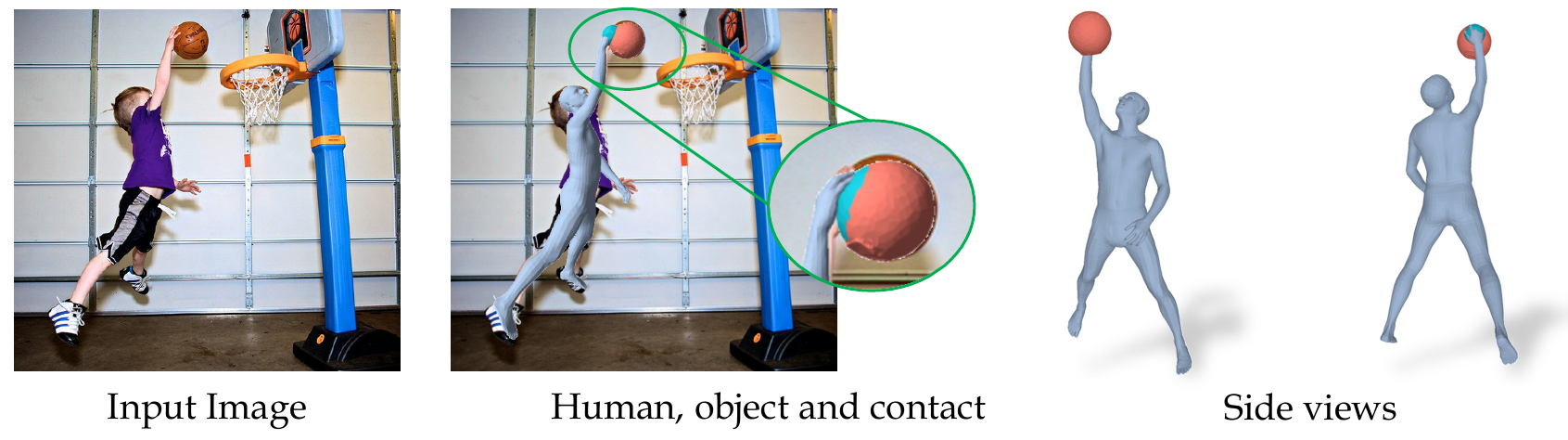}
    \caption{From a single image, our method jointly reconstructs the 3D human pose and shape coherently with the object by reasoning about their interactions. Instead of predicting human and object independently and imposing coherency post-hoc\cite{zhang2020phosa}, our method learns to reconstruct both simultaneously including contacts, leading to significant improvements over the state-of-the-art.}
    \label{fig:teaser}
\end{figure}
In order to deploy robots and intelligent systems in the real world, they must be able to perceive and understand humans interacting with the real world from visual input.  
While there exists a vast literature in perceiving humans in 3D from single images, the large majority of works perceive humans in isolation~\cite{kanazawa2018endtoend,omran2018neural,alldieck2019tex2shape,saito2020pifuhd,VIBE:CVPR:2020, xiu2022icon}. The joint 3D perception of humans and objects has received much less attention and is the main focus of this work. Joint reconstruction of humans and objects is extremely challenging. The object and human occlude each other making inference hard; it is difficult to predict their relative size and spatial arrangement in the 3D world, and the visual evidence for contacts and interactions consists of only a very small region in the image.

Recent work~\cite{zhang2020phosa} addresses these challenges by reconstructing 3D objects and humans separately, and imposing hand-crafted rules like manually defined object contact regions and object size. However, heuristics cannot scale as it is difficult to specify all possible human-object contacts beforehand. Consider the chair in ~\cref{fig:orth_vs_persp}: one can sit, lean, move it from the handle, back or legs, which constitutes different pairs of human-object contacts that are not easy to manually enumerate. Furthermore, reasoning human and object separately can produce errors that cannot be corrected afterwards, see \cref{fig:phosa_comparison}. These motivate a method that can jointly reconstruct humans and objects and learn the interactions from data, without hand-crafted rules.

In this work, we introduce CHORE, a novel method to jointly recover a 3D parametric body model and the 3D object, from a single image.
CHORE is built on two high-level ideas. First, instead of reconstructing human and object separately, we argue for jointly reason them in the first place. Secondly, we learn spatial configuration priors of interacting human and object from data without using heuristics. 
Instead of regressing human parameters directly from image, we first obtain neural distance fields to human and object surfaces as well as a correspondence field to the SMPL~\cite{smpl2015loper} body model. To obtain robust object fitting, we also predict a neural field for object rotation and translation. This allows us to formulate a more robust optimization energy to fit the SMPL model and the object template to the image. To train our network on real data, we propose a training strategy that enables us to effectively learn pixel-aligned implicit functions with perspective cameras, which leads to more accurate prediction and joint fitting. 
In summary, our key contributions are:
\begin{itemize}
    \item We propose CHORE, the first end-to-end learning approach that can reconstruct human, object and contacts from a single RGB image. Our correspondence and contact prediction allow us to accurately register a controllable body model and 3D object to the image.
    \item Different from prior works that use weak perspective cameras and learn from synthetic data, we use perspective camera model, which is crucial to train on real data. To this end, we propose a new training strategy that allows effective pixel-aligned implicit learning with perspective cameras. 
    \item Through our effective training and joint reconstruction, our model achieves a remarkable performance improvement of over $50\%$  compared to prior art \cite{zhang2020phosa}. Our code and models are publicly available to foster future research. 
\end{itemize}

\section{Related Work}
We propose the first learning-based approach to jointly reconstruct 3D human, object and contacts from a single image. In this regard, we first review recent advances in separate human or object reconstructions. We then discuss approaches that can jointly reason about human-object interaction.

\textbf{Human reconstruction from images.} One common approach to reconstruct 3D humans from images is to fit a parametric model such as SMPL~\cite{smpl2015loper} to the image~\cite{bogo2016smplify,SMPL-X:2019,alldieck2018video,alldieck2018detailed,kolotouros2019spin, coronaLVD}. Learning-based approaches have also been used to directly regress model parameters such as pose and shape~\cite{omran2018neural,  kanazawa2018endtoend,pavlakos2018humanshape,VIBE:CVPR:2020,Guler_2019_CVPR, jiang2020mpshape} as well as clothing~\cite{bhatnagar2019mgn,alldieck2019learning,jiang2020bcnet}.
More recently, implicit function based reconstruction are applied to capture fine-grained details including clothing and hair geometry~\cite{pifuSHNMKL19,Huang2020ARCHAR,saito2020pifuhd,ZHANG20211, xiu2022icon}. These methods however cannot reconstruct 3D objects, let alone human-object interaction.

\textbf{Object reconstruction from images.}
Given a single image, 3D objects can be reconstructed as voxels~\cite{3dr2n2,marrnet,HierarchicalSurfac2017hane}, point clouds~\cite{Fan_2017_CVPR,pix2surf_2020} and meshes~\cite{Kar_category-specificobject,Wang_2018_ECCV,image2mesh}.
Similar to human reconstruction, implicit functions have shown huge success for object reconstruction as well~\cite{Mescheder2019OccNet,DISN,mueller2021completetracking}.
The research in this direction has been significantly aided by large scale datasets ~\cite{shapenet2015,objectnet3d2016yu,Sun_2018_CVPR}. We urge the readers to see the recent review~\cite{review2021reconstruction} of image-based 3D reconstruction of objects.
These approaches have shown great promise in rigid object reconstruction but articulated humans are significantly more challenging and human-object interaction even more so. Unlike these methods, our approach can jointly reconstruct humans, objects and their interactions.

\textbf{Human object interactions.}
Modelling 3D human-object interactions is extremely challenging. Recent works have shown impressive performance in modelling hand-object interactions from 3D~\cite{GRAB:2020,ContactGrasp2019Brahmbhatt, zhou2022toch}, 2.5D~\cite{Brahmbhatt_2019_CVPR,Brahmbhatt_2020_ECCV} and images~\cite{GrapingField:3DV:2020,Corona_2020_CVPR,hasson19_obman, ehsani2020force, yang2021cpf}.
Despite being quite impressive, these works have a shortcoming in that they are restricted to only hand-object interactions and cannot predict the full body.
Modelling full-body interactions is even more challenging, with works like PROX~\cite{PROX:2019} and \cite{weng2020holistic} being able to fit 3D human model to satisfy the 3D scene constraints~\cite{savva2016pigraphs,PLACE:3DV:2020,Hassan:CVPR:2021, zhang2022couch}, capture interaction from mult-view~\cite{behavecvpr22, sun2021HOI-FVV, jiang2022neuralfusion} or reconstruct 3D scene based on human-scene interactions~\cite{Yi_MOVER_2022}. 
Recently, this direction has also been extended to model human-human interaction~\cite{Fieraru_2020_CVPR},  self-contacts~\cite{selfcontact2020Fieraru, Mueller:CVPR:2021:selfcontact}.
Broadly speaking, these works can model human-object interactions to to satisfy scene constraints but cannot jointly reconstruct human-object contacts from single images. Few works directly reason about 3D contacts from images~\cite{chen2019holisticpp,caoHMP2020, huang2022rich} and video~\cite{RempeContactDynamics2020,Li_2019_CVPR,MonszpartEtAl:iMapper:Siggraph:2019} but do not reconstruct 3D human and object from a single RGB image.

Closest to our work are \cite{weng2020holistic} and PHOSA~\cite{zhang2020phosa} . Weng \etal \cite{weng2020holistic} predict 3D human and scene layout separately and use scene constraints to optimize human reconstruction. They use predefined contact weights on SMPL-X\cite{SMPL-X:2019} vertices to optimize human meshes, which is prone to errors. PHOSA first fits SMPL and object meshes separately and then use hand crafted heuristics like predefined contact pairs to reason human object interaction. These heuristics are neither scale-able nor very accurate.
On the other hand, our method does not rely on heuristics and can learn joint reconstruction and contact priors directly from data. Our experiments show that our learning based method easily outperforms heuristic-based PHOSA and \cite{weng2020holistic}.

\section{Method}
\begin{figure*}[t]
    \centering
    \includegraphics[width=\linewidth]{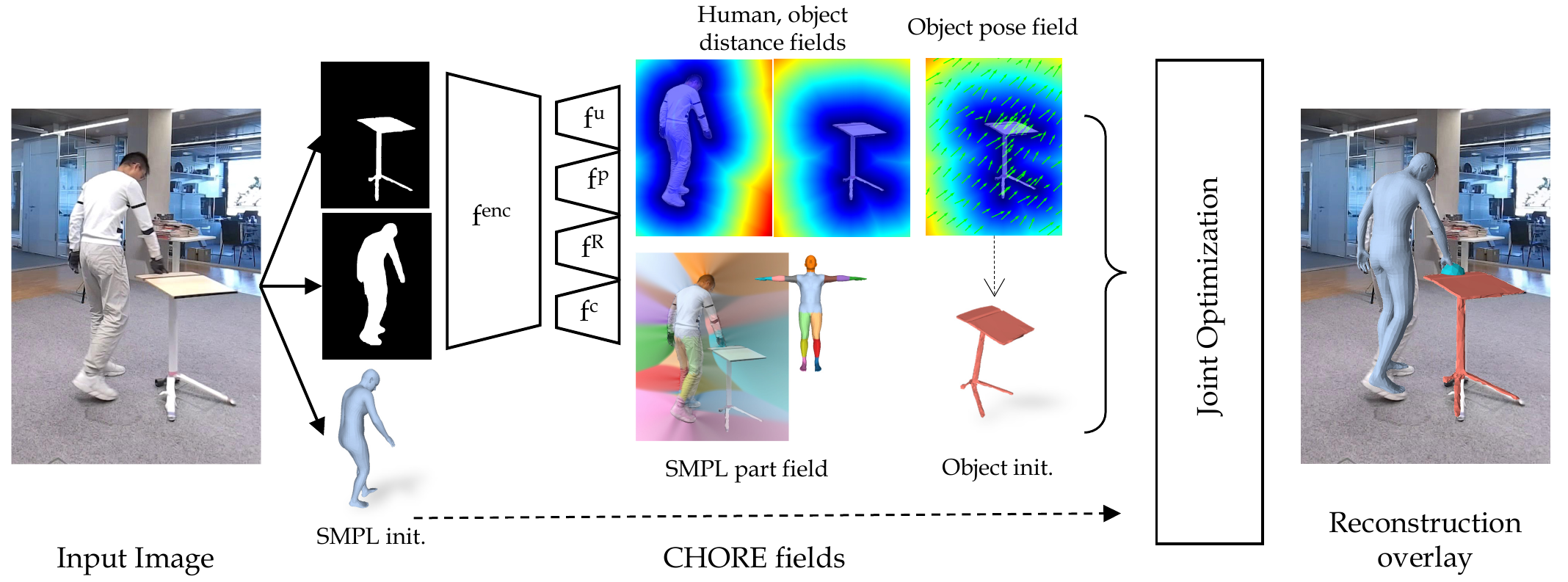}
    \caption{We present an approach to jointly reconstruct human, object and contacts from a single RGB image. Given an image and human, object masks, we predict CHORE fields: unsigned distance fields to human and object surfaces ($f^u$), SMPL part ($f^p$) and object pose ($f^R, f^c$) fields. We then leverage the neural fields to robustly fit the SMPL model and the object mesh to the image while reasoning human-object contacts.
    }
    \label{fig:overview}
\end{figure*}
We describe here the components of CHORE, a method for 3D reconstruction of the human, object and their contacts from a single RGB image. This work focuses on a single person in close interaction with one dynamic movable object.
Our approach is inspired by the classical image based model fitting methods \cite{alldieck2018video,bogo2016smplify}, which fit a body model parameterized by shape and pose $\mathbf{\pose,\shape}$ by minimizing an energy: 
\begin{equation}
    E(\mathbf{\pose, \shape})=E_{data} + E_{J2D} +E_{reg}, 
    \label{eq:fit_human}
\end{equation} 
where $E_{data}$ is a 2D image based loss like a silhouette loss, $E_{J2D}$ is re-projection loss for body joints, and $E_{reg}$ is a body pose and shape prior loss. Estimating the human and the object jointly however is significantly more challenging. The relative configuration of human and object needs to be accurately estimated despite occlusions, lack of depth, and hard to detect contacts. Hence, a plain extension of a hand-crafted 2D objective to incorporate the object is far too prone to local minima.
Our idea is to jointly learn to predict, based on a single image, 3D neural fields (CHORE fields) for the object, human and their relative pose and contacts, to formulate a learned robust 3D data term.  
We then fit a parametric body model and a template object mesh to the predicted CHORE neural fields, see overview \cref{fig:overview}.

In this section, we first explain our CHORE neural fields in \cref{sec:neural_recon}, and then describe our novel human-object fitting formulation using learned CHORE neural fields in \cref{sec:joint_fitting}. 
Unlike prior work on pixel-aligned neural implicit reconstruction (PiFu~\cite{pifuSHNMKL19}) which learns from synthetic data and assume a weak camera model, we learn from real data captured using perspective cameras. This poses new challenges which we address with a simple and effective deformation of the captured real 3D surfaces to account for perspective effects (\cref{sec:centering}). 

\subsection{CHORE neural fields from single images}\label{sec:neural_recon} 
As representation, we use the SMPL body model $\smpl(\pose, \shape)$~\cite{smpl2015loper} which parameterizes the 3D human as a function of pose $\pose$ (including global translation) and shape $\shape$, and a template mesh for the object. 
Our idea is to leverage a stronger 3D data term coming from learned CHORE neural fields  to fit SMPL and the object template to the image.
In contrast to a plain joint surface reconstruction of human and object, the resulting SMPL and object mesh can be further controlled and edited. 
To obtain a good fit to the image, we would ideally i) minimize the distance between the optimized meshes and the corresponding ground truth human and object surfaces, and ii) enforce that human-object contacts of the meshes are consistent with input. 
Obviously, ground truth human and object surfaces are unknown at test time. Hence, we approximate them with CHORE fields. Specifically, a learned neural model predicts 1) the unsigned distance fields of the object and human surfaces, and 2) a part correspondence field to the SMPL model, for more robust SMPL fitting and contact modeling, and 3) an object pose field, which we use to initialize the object pose.

This constitutes a complex multi-task learning problem which we address with a single neural network. 
We first use a CNN to extract a image feature grid. For a query point $\vect{p}\in \mathbb{R}^3$ we project it to image plane and extract pixel-aligned features. From these features we predict the distance, part-correspondence and object pose fields. We explain the details of our network below.

\textbf{Input encoding $f^{enc}$.} We stack RGB image with human and object masks in separate channels as input and use a leaned feature extractor $f^{enc}$ \cite{pifuSHNMKL19}to extract image feature grid $\mat{F}$. 
Given a query point $\vect{p} \in \mathbb{R}^3$, and the image feature $\mat{F}$, we extract a pixel-aligned point feature $\mat{F}[\pi_{3D \mapsto 2D}(\vect{p})]$ by projecting the 3D point to the image. 
We concatenate the 3D point coordinate $\vect{p}$ to the pixel-aligned feature, allowing the network to distinguish points at different depth along the same ray,
hence $\mat{F}_\vect{p} = (\mat{F}[\pi_{3D \mapsto 2D}(\vect{p})],\vect{p})$. The feature extractor $f^{enc}$ is trained with losses defined on neural fields predictions which we explain later. 

\textbf{Human and object distance fields $f^u$.}\label{sec:surface_recon} Based on the point feature $\mat{F}_\vect{p}$, we use a point decoder $f^u$ to predict the unsigned distances $u_h(\vect{p})\in\mathbb{R}^+$ and $u_o(\vect{p})\in\mathbb{R}^+$ from the point to the human and object surface respectively. This allows to model non-watertight meshes as well. 
From an unsigned distance field the 3D surface $\set{S}$ can be reconstructed by projecting points to it~\cite{chibane2020ndf}:
\begin{equation}
    \set{S}_i = \{\vect{p} - f^u_i(\mat{F}_\vect{p}) \nabla_\vect{p} f^u_i(\mat{F}_\vect{p})\, | \, \vect{p} \in \mathbb{R}^3\}
    \label{eq:udf_recon}
\end{equation}where $f^u_i$ can be human ($f^u_h$) or object ($f^u_o$) UDF predictor. In practice, points $\vect{p}$ are initialized from a fixed 3D volume and projection is done iteratively\cite{chibane2020ndf}. 

To train neural distance function $f^u_i$, we sample a set of points $\mathcal{P}_i$ near the ground truth surfaces and compute the ground truth distance $\text{UDF}_\text{gt}(\vect{p})$. The  decoder $f^u_i$ is then trained to minimize the $L_1$ distance between clamped prediction and ground truth distance:
\begin{equation}
    L_{u_i} = \sum_{\vect{p}\in \mathcal{P}_i} |\min(f^u_i(\mat{F}_{\vect{p}}), \delta) - \min(\text{UDF}_\text{gt}(\vect{p}), \delta)|
\end{equation}
where $\delta$ is a small clamping value to focus on the volume nearby the surface. 

\textbf{Part field $f^p$ and contacts.} \label{sec:part_pred}
While the unsigned distance predictions can reconstruct the human and object surface, they do not tell which human or object points are in contact, and fitting template SMPL meshes using only distance fields often fails~\cite{bhatnagar2020ipnet}. Hence we use another decoder to predict a part correspondence field to the SMPL model with two purposes: i) to formulate a robust SMPL fitting objective~\cite{bhatnagar2020ipnet, bhatnagar2020loopreg} and ii) to derive which body parts are in contact with the object points. Mathematically, given the point feature $\mat{F}_\vect{p}$ for a point $\vect{p} \in \mathbb{R}^3$, we predict its corresponding SMPL part $\vect{l}_\vect{p}=f^p(\mat{F}_\vect{p}) \in \{1,2,...K\}$. $K=14$ is the total number of SMPL parts. The part prediction decoder is trained with standard categorical cross entropy, $L_p$.

\textbf{Object pose fields $f^R$ and $f^c$.}\label{sec:object_pose}
Accurate object fitting to the CHORE UDF requires good initialization for the object pose. 
We learn a predictor $f^R: \mat{F_\vect{p}}\mapsto \mathbb{R}^{3\times3}$ that takes the point feature $\mat{F}_\vect{p}$ as input and predicts the rotation matrix $\mat{R}$. 
Predicting the object translation in camera coordinates directly is more tricky due to the well known depth-scale ambiguity. Hence our predictions are relative to the human.
We train a decoder $f^c: \mat{F_\vect{p}}\mapsto \mathbb{R}^5$ which predicts the x, y coordinate of SMPL center (at fixed depth) in camera coordinates, and the object center relative to the SMPL center based on the point feature $\mat{F}_\vect{p}$. Both $f^R$ and $f^c$ are trained with mean squared loss $L_R$ and $L_c$. 

Our feature encoder $f^{enc}$ and neural predictors $f^u, f^p, f^R, f^c$ are trained jointly with the objective: $L = \lambda_u (L_{u_h}+L_{u_o}) + \lambda_p L_p + \lambda_R L_R + \lambda_c L_c$. 
See supplementary for more details on our network architecture and training.
\subsection{3D human and object fitting with CHORE fields}\label{sec:joint_fitting}
With CHORE fields, we reformulate \cref{eq:fit_human} to include a stronger data term  $E_{data}$ in 3D, itself composed of 3 different terms for human, object, and contacts which we describe next.

\textbf{3D human term.} This term consists of the sum of distances from SMPL vertices to the CHORE neural UDF for the human $f^u_h$, and a part based term leveraging the part fields $f^p$ discussed in \cref{sec:surface_recon}:
\begin{equation}
\label{eq:our_objective_human_label}
   E^h_\text{data}(\pose, \shape) = \sum_{\vect{p}\in \smpl(\pose, \shape)}(\lambda_{h}\min(f^u_h(\mat{F}_\vect{p}), \delta) + \lambda_{p\prime} L_p(l_\vect{p}, f^p(\mat{F}_\vect{p}))),
\end{equation}
Here, $L_p(\cdot, \cdot)$ is the standard categorical cross entropy loss function and $l_{\vect{p}}$ denotes our predefined part label on SMPL vertex $\vect{p}$. $\lambda_{h}, \lambda_{p\prime}$ are the loss weights. 
The part-based term $L_p(\cdot, \cdot)$ is effective to avoid, for example, matching a person's arm to the torso or a hand to the head. 
To favour convergence, we initialize the human pose estimated from the image using \cite{rong2020frankmocap}. 

\textbf{3D object term.} We assume a known object mesh template with vertices $\mat{O}\in \mathbb{R}^{3\times N}$
and optimize the object rotation $\mat{R}_o\in SO(3)$, translation $\vect{t}_o\in \mathbb{R}^3$ and scale $s_o\in \mathbb{R}$ leveraging the neural fields predictions. Given template vertices and object pose, we compute the posed object as $\mat{O}^\prime=s_o(\mat{R}_o\mat{O} + \vect{t}_o)$. 
Intuitively, the distance between object vertices $\mat{O}^\prime$ and underlying object surface should be minimized. Furthermore, the projected 2D silhouette should match the observed object mask $\mat{M}_o$,  yielding the following objective: 
\begin{equation}
\label{eq:our_objective_object}
    E_\text{data}^o(\mat{R}_o, \vect{t}_o, s_o) = \sum_{\vect{p}\in \mat{O}^\prime}(\lambda_o\min(f^u_o(\mat{F}_\vect{p}), \delta)  + \lambda_\text{occ}L_\text{occ-sil}(\mat{O}^\prime, \mat{M}_o) + \lambda_\text{reg}L_\text{reg}(\mat{O}^\prime)),
\end{equation} 
The first term computes the distance between fitted object and corresponding surface represented implicitly with the network $f_o^u$. The second term $L_\text{occ-sil}$ is an occlusion-aware silhouette loss \cite{zhang2020phosa}. We additionally use $L_\text{reg}$ computed as the distance between predicted object center and center of $\mat{O}^\prime$. This regularization prevents the object from being pushed too far away from predicted object center. 

The object rotation is initialized by first averaging the rotation matrices of the object pose field near the object surface (points with $f^u_o(\mat{F}_\vect{p})<4mm$): $\mat{R}_o=\frac{1}{M}\sum^M_{k=1}\mat{R}_k$. We further run SVD on $\mat{R}_o$ to project the matrix to $SO(3)$.
Object translation $\vect{t}_o$ is similarly initialized by averaging the translation part of the pose field nearby the surface. 
Our experiments show that pose initialization is important for accurate object fitting, see \cref{sec:ablate-object-center}.

\textbf{Joint fitting with contacts.} While \cref{eq:our_objective_human_label} and \cref{eq:our_objective_object} allow fitting human and object meshes coherently,
fits will not necessarily satisfy fine-grained contact constraints. Our key idea here is to jointly reconstruct and fit humans and objects while reasoning about contacts. 
We minimize a joint objective as follows:
\begin{equation}
\label{eq:final_objective}
    E_\text{data}(\pose, \shape, \mat{R_o}, \vect{t_o}, s_o) = E_\text{data}^h + E_\text{data}^o + \lambda_cE_\text{data}^c,
\end{equation}
where $E_\text{data}^h$ and $E_\text{data}^o$ are defined in \cref{eq:our_objective_human_label} and \cref{eq:our_objective_object} respectively. The contact term $E_\text{data}^c$ consists of the chamfer distance $d(\cdot,\cdot)$ between the sets of human $\smpl_j^c$ and object points $O_j^c$ predicted to be in contact:
\begin{equation}
\label{eq:our_objective_contact}
    E_\text{data}^c(\mat{R}_o, \vect{t}_o, s_o) = \sum_{j=1}^{K} d(\smpl_j^c(\pose, \shape), O_j^c).
\end{equation}
Points on the $j-th$ body part of SMPL ($H_j(\pose, \shape)$) are considered to be in contact when their UDF $f^u_o(\mat{F}_\vect{p})$ to the object is smaller than a threshold: $\smpl_j^c(\pose, \shape)=\{\vect{p}\,|\,\vect{p}\in \smpl_j(\pose, \shape) \, \textrm{and} \, f^u_o(\mat{F}_\vect{p})\leq\epsilon \}$.
Analogously, object points are considered to be in contact with body part $j$ when their UDF to the body is small and they are labelled as part $j$: $O_j^c=\{\vect{p}\,|\,\vect{p}\in \mat{O}^\prime \, \textrm{and} \, f^u_h(\mat{F}_\vect{p})\leq \epsilon \  \textrm{and} \, f^p(\mat{F}_\vect{p})=j\}$.
The contact term $E_\text{data}^c$ encourages the object to be close with the corresponding SMPL parts in contact. The loss is zero when no contacts detected. This is important to have a physically plausible and consistent reconstruction, see \cref{sec:contact}.

Substituting our data term into \cref{eq:fit_human}, we can now write our complete optimization energy as:    $E(\pose, \shape, \mat{R_o}, \vect{t_o}, s_o)=E_{data}^o + E_{data}^h +\lambda_cE_{data}^c+\lambda_\text{J}E_{J2D} +\lambda_rE_\text{reg}$. See Supp. for details about different loss weights $\lambda$'s. 

\subsection{Learning pixel-aligned neural fields on real data}\label{sec:centering}
\begin{figure}[t]
    \centering
    \includegraphics[width=\linewidth]{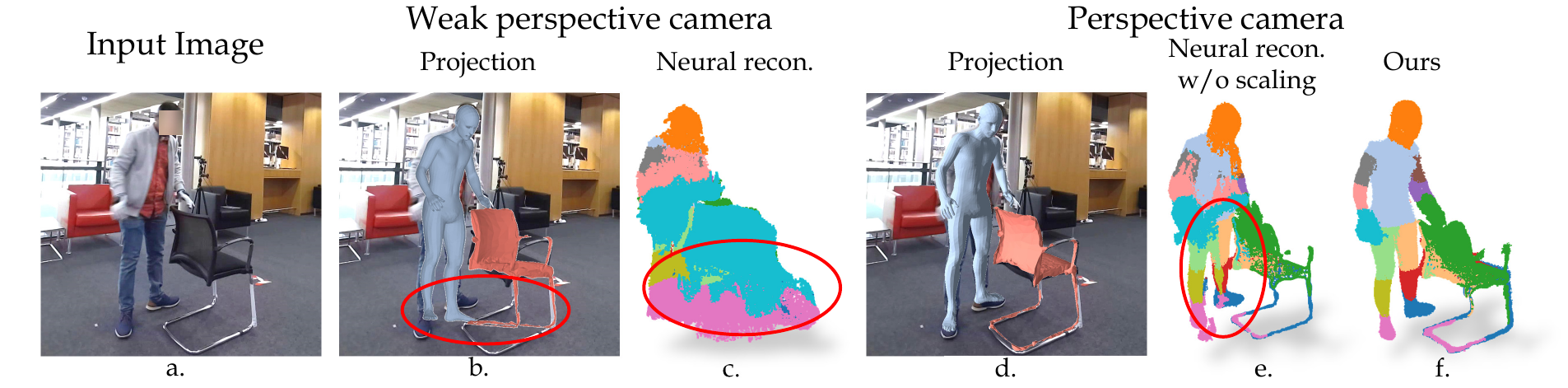}
    \caption{Learning neural fields on real data. We show projections of GT meshes overlapped with input image using different camera models in b and d. The neural reconstructions (colored based on predicted part labels $f^p$) are shown in c, e and f. Weak perspective projection is not accurately aligned (b) with image hence network trained with this predicts noisy output (c). Perspective camera projection is accurately aligned (d) but directly applying it, without accounting for depth-scale ambiguity, makes learning hard. Training with this is still not optimal (e). Our proposed method reduces depth ambiguity while maintaining accurate alignment, leading to better predictions (f). }
     \label{fig:orth_vs_persp} 
\end{figure}
Our training data consists of real images paired with reference SMPL and object meshes.
Here, we want to follow the pixel-aligned training paradigm, which has been shown effective for 3D shape reconstruction. 
Since both depth and scale determine the size of objects in images this makes learning from single images ambiguous. 
Hence, existing works place the objects at a fixed distance to the camera, and render them using a weak perspective~\cite{pifuSHNMKL19, zhao2021learning-anchor, saito2020pifuhd} or perspective camera model~\cite{genre, shapehd, marrnet, DISN}. 
This effectively allows the network to reason about only scale, instead of both scale and depth at the same time. 

This strategy to generate training pairs is only possible when learning from \emph{synthetic data}, which allows to move objects and re-render the images.
Instead, we learn from \emph{real} images paired with meshes. Re-centering meshes to a fixed depth breaks the pixel alignment with the real images. 
An alternative is to adopt a weak perspective camera model (assume all meshes are centered at a constant depth), but the alignment is not accurate hence leads to inaccurate learning, see \cref{fig:orth_vs_persp} b) and c). 
Alternatively, attempting to learn a model directly from the original pixel-aligned data with the intrinsic depth-scale ambiguity leads to  poor results, as we show in \cref{fig:orth_vs_persp} e). 
Hence, our idea is to adopt a perspective camera model, and transform the data such that it is centered at a fixed distance to the camera, while preserving pixel alignment with the real images.
We demonstrate that this can be effectively achieved by applying depth-dependent scale factor. 

Our first observation is that scaling the mesh vertices $\mat{V}$ by $s$ does not change its projection to the image. \emph{Proof:} Let $\vect{v}=(\mat{v}_x,\mat{v}_y,\mat{v}_z)\in\mat{V}$ be a vertex of the original mesh, and $\vect{v}^\prime=(s\mat{v}_x,s\mat{v}_y,s\mat{v}_z) \in s\mat{V}$ a vertex of the scaled mesh.
For a camera with perspective focal length $f_x, f_y $ and camera center $c_x, c_y$, the vertex $\vect{v}$ is projected to the image as $\pi(\vect{v})=(f_x\frac{\vect{v}_x}{\vect{v}_z}+c_x, f_y\frac{\vect{v}_y}{\vect{v}_z}+c_y)$. The scaled vertices will be projected to the exact same pixels, $\pi(\vect{v}^\prime)=(f_x\frac{s\vect{v}_x}{s\vect{v}_z}+c_x, f_y\frac{s\vect{v}_y}{s\vect{v}_z}+c_y) = \pi(\vect{v})$ because the scale cancels out. This result might appear un-intuitive at first. What happens is that the object size gets scaled, but its center is also scaled, hence making the object bigger/smaller and pushing it further/closer to the camera at the same time. This effects cancel out to produce the same projection. 

Given this result, all that remains to be done, is to find a scale factor which centers objects at a \emph{fixed} distance $z_0$ to the camera. 
The desired scale factor that satisfies this condition is $s=\frac{z_0}{\mu_z}$, where $\mu_z = \frac{1}{n} \sum_{i=1}^n \vect{v}^i_z$ is the mean of the un-scaled mesh vertices depth. 
\emph{Proof:} The proof follows immediately from the linearity property of the empirical mean. 
The depth of the transformed mesh center is: 
\begin{equation}
    \mu_{z}^\prime=\frac{1}{n}\sum^n_{i=1}  \vect{v}_z^{i\prime}=\frac{1}{n}\sum^n_i \vect{v}_z^{i}\frac{z_0}{\mu_z} = \mu_z \frac{z_0}{\mu_z} = z_0
\end{equation} 
This shows that the depth $\mu_z^\prime$ of the scaled mesh center is fixed at $z_0$ and its image projection remains unchanged, as desired.

We scale all meshes in our training set as described above to center them at $z_0=2.2m$ and recompute GT labels. To maintain pixel alignment we use the camera intrinsic from the dataset.
At test time, the images have various resolution and \textit{unknown} intrinsic with person at large range of depth (see \cref{fig:coco-example}). We hence crop and resize the human-object patch such that the person in the resized patch appears as if they are at $z_0$ under the camera intrinsic we use during training. We leverage the human mesh reconstructed from \cite{rong2020frankmocap} as a reference to determine the patch resizing factor. Please see supplementary for details.

\section{Experiments}
\begin{figure*}[t]
    \centering
    \includegraphics[width=\linewidth]{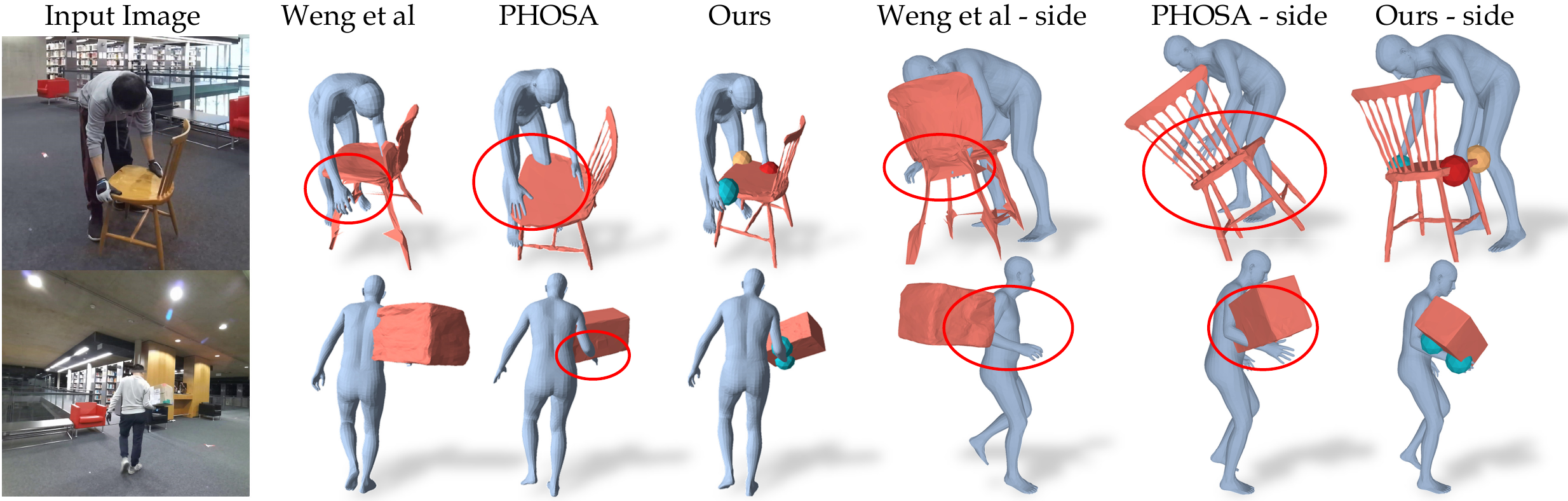}
    \caption{Comparison with Weng \etal~\cite{weng2020holistic} and PHOSA~\cite{zhang2020phosa} on \behave{}. The contact spheres are placed automatically at the contact centers.  \cite{weng2020holistic} uses predefined contact weights and fails to reconstruct spatial arrangement properly. PHOSA's post-hoc also fails to correct errors from separate human-object reconstructions. Our network is trained jointly to reconstruct the human and the object and learns a strong prior on the relative arrangement, allowing for more accurate reconstruction.}
    \label{fig:phosa_comparison}
\end{figure*}

\textbf{Datasets.} We conducted experiments on the \behave{} \cite{behavecvpr22}, COCO \cite{coco-dataset} and NTU-RGBD\cite{Liu_2019_NTURGBD120} dataset. \behave{} captures 8 subjects interacting with 20 different objects in natural environments using RGBD cameras. Each image is paired with (pseudo) ground truth SMPL and object fits in camera coordinate.  CHORE is trained on 10.7k frames at 4 locations from \behave{} and tested on 4.5k frames from one unseen location.  All 20
categories are seen in both training and test set but at test time the object instances maybe different. 
As PHOSA~\cite{zhang2020phosa} relies on 2D object mask for pose estimation, we exclude images where the object is occluded more than 70\% for fair comparison, which leads to 4.1k test frames. To compare with Weng \etal~\cite{weng2020holistic}, we select 10 common object categories between \behave{} and their method (in total 2.6k images) and compare our method with them on this smaller set, denoted as $\mathrm{\behave{}^{--}}$.

To verify the generalization ability, we also tested our method on COCO~\cite{coco-dataset}  where each image contains at least one instance of the object categories in \behave{} and the object should not be occluded more than 50\%. Images where there is no visible contact between human and object are also excluded. There are four overlapping categories between \behave{} and COCO (backpack, suitcase, chair and sports ball) and in total 913 images are tested.  

We additionally test on the NTU-RGBD dataset, which is quite interesting since it features human-object interactions. We select images from the NTU-RGBD dataset following the same criteria for COCO and in total 1.5k images of chair or backpack interactions are tested.

\textbf{Evaluation and baselines.} To the best of our knowledge, PHOSA \cite{zhang2020phosa} and Weng \etal \cite{weng2020holistic} are the closest works. 
To compare with PHOSA~\cite{zhang2020phosa}, we annotated template object meshes with contact part labels and intrinsic scales according to PHOSA's convention. 
\cite{weng2020holistic} focuses on reconstructing the human in static scenes, hence their objects are mainly from indoor environments. Therefore, we only compare with it on the $\mathrm{\behave{}^{--}}$ set and the indoor NTU-RGBD dataset. 

For \behave{}, we compute the Chamfer distance separately on SMPL and object meshes after Procrustes alignment using the combined mesh. Since \cite{weng2020holistic} uses a different object template, we perform Procrustes alignment using only SMPL vertices for $\mathrm{\behave{}^{--}}$. As there are no 3D ground truth available in COCO and NTU-RGBD for thorough quantitative comparison, we compare our method with baselines through anonymous user study surveys.

\subsection{Comparison on \behave{} dataset}
\begin{figure}[t]
    \centering
    \includegraphics[width=\linewidth]{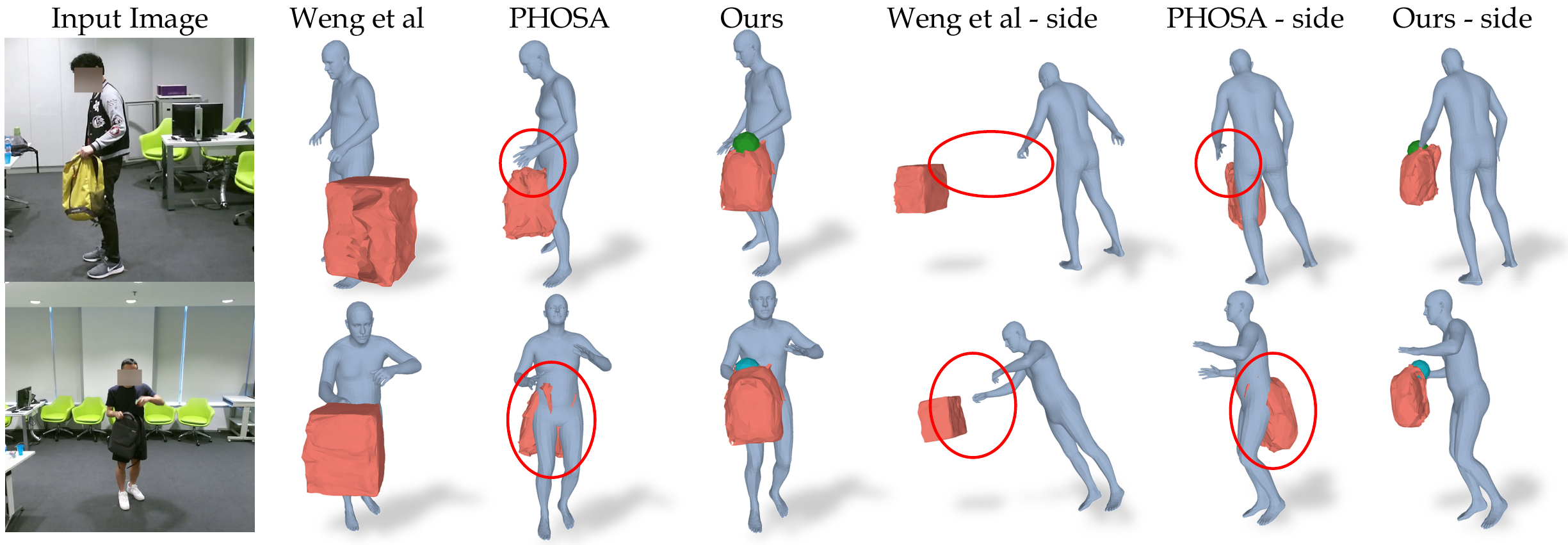}
    \caption{Comparing our methods with \cite{zhang2020phosa} and \cite{weng2020holistic} on NTU-RGBD dataset\cite{Liu_2019_NTURGBD120}.  Our method trained on \behave{} generalizes to NTU-RGBD and outperforms baselines.}
    \label{fig:comp_ntu}
\end{figure}
We compare our method against PHOSA \cite{zhang2020phosa} and \cite{weng2020holistic} in~\cref{table:phosa}. Note that the Procrustes alignment is performed on combined mesh for \behave{} and on SMPL mesh only for $\mathrm{\behave{}^{--}}$ for a fair comparison with \cite{weng2020holistic}. 
~\cref{table:phosa} clearly shows that our learning based approach outperforms the heuristic optimization based PHOSA and \cite{weng2020holistic}.
\begin{table}[h]
    \centering
    \begin{tabular}{l | c c c c}
    \toprule[1.5pt]
    {\bf Dataset} & {\bf Methods} & {\bf SMPL $\downarrow$} & {\bf Object $\downarrow$} \\
    \midrule
    \multirow{2}{*}{\behave{}} &PHOSA \cite{zhang2020phosa}& 12.17 $\pm$ 11.13 & 26.62 $\pm$ 21.87\\
    & { Ours} & {\bf 5.58 $\pm$ 2.11} & {\bf 10.66 $\pm$ 7.71}\\
    \midrule
    \multirow{3}{*}{$\mathrm{\behave{}^{--}}$} &PHOSA \cite{zhang2020phosa} & 6.09 $\pm$ 1.98 & 182.67 $\pm$ 319.64\\
    & Weng et al \cite{weng2020holistic} & { 7.86 $\pm$ 3.48} & {51.86 $\pm$ 96.38} \\
    & { Ours} & {\bf 5.22 $\pm$ 2.03} & {\bf 13.15 $\pm$ 10.60}\\
    \bottomrule[1.5pt]
    \end{tabular}
    \caption{The mean and standard deviation of the Chamfer distance (cm) for SMPL and object reconstruction on \cite{behavecvpr22}. Our approach that jointly reasons about the human, object and the contacts outperforms \cite{weng2020holistic} and heuristics based PHOSA~\cite{zhang2020phosa}. }
    \label{table:phosa}
\end{table}

We also compare our method qualitatively with PHOSA and \cite{weng2020holistic} in~\cref{fig:phosa_comparison}. It can be seen that PHOSA fails to reconstruct the object at the correct position relative to the human. \cite{weng2020holistic} also fails quite often as it uses predefined contact weights on SMPL vertices instead of predicting contacts from inputs. On the other hand, CHORE works well because it learns a 3D prior over human-object configurations based on the input image. See supplementary for more examples.

\subsection{Generalisation beyond \behave{}}
\begin{figure*}[t]
    \centering
    \includegraphics[width=1.0\linewidth]{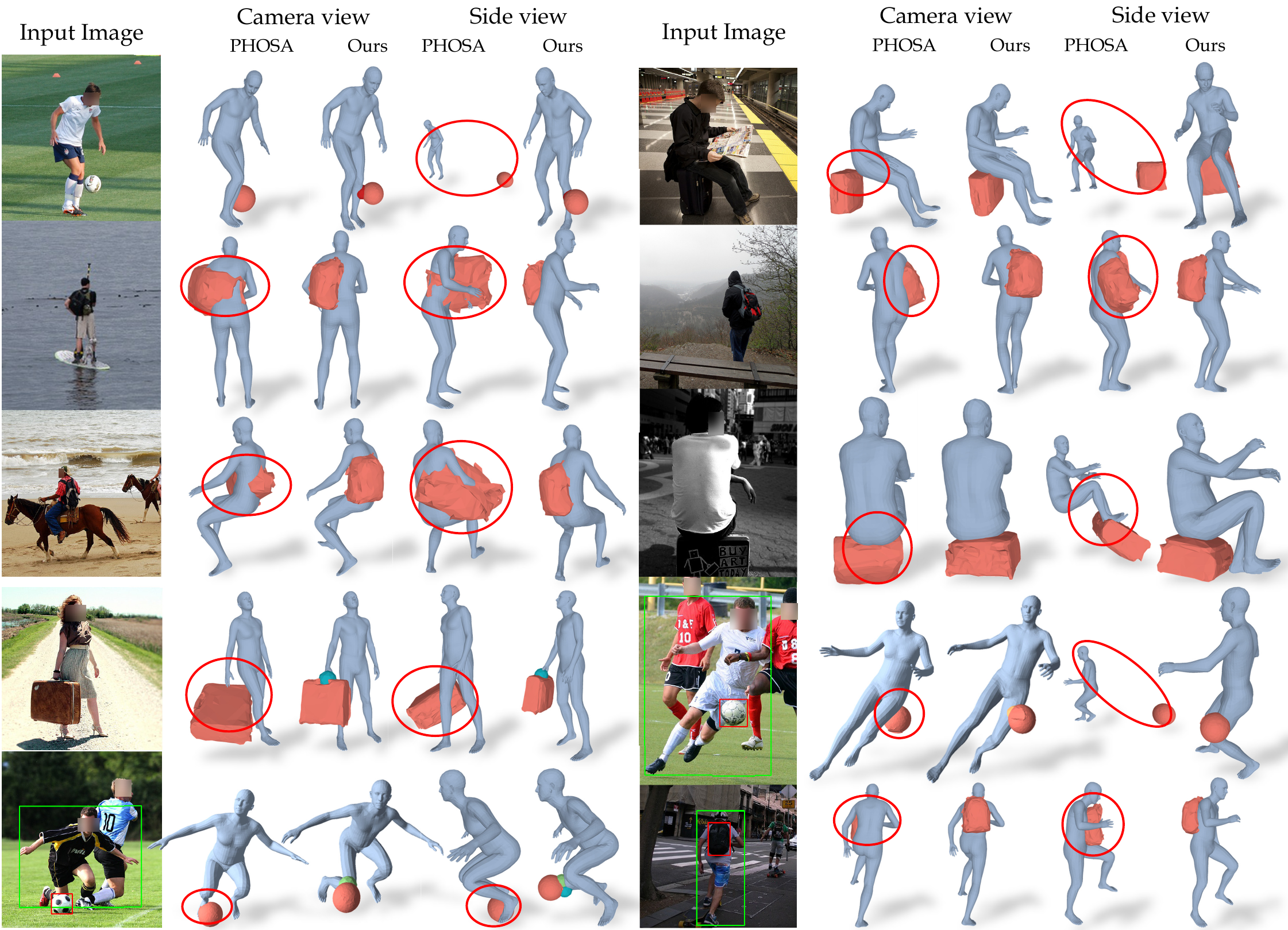}
    \caption{Comparison with PHOSA\cite{zhang2020phosa} on COCO dataset\cite{coco-dataset}. For images with more than one person, the reconstructed human and object are highlighted by green and red boxes. Our model trained on \behave{} outperforms PHOSA on in the wild images.}
    \label{fig:coco-example}
\end{figure*}

To verify our model trained on the \behave{} dataset can generalize to other datasets, we further test it on the NTU-RGBD \cite{Liu_2019_NTURGBD120} as well as COCO dataset \cite{coco-dataset} and compare with \cite{zhang2020phosa, weng2020holistic}. 

\textbf{Comparison with PHOSA\cite{zhang2020phosa}.}
We test our method and PHOSA on the selected COCO images discussed above. We then \emph{randomly} select 50 images and render ours and PHOSA reconstruction results side by side. The results are collected in an anonymous user study survey and we ask 50 people in Amazon Mechanical Turk (AMT) to select which reconstruction is better. Notably, people think our reconstruction is better than PHOSA on 72\% of the images. A similar user study for images from the NTU-RGBD dataset is released to AMT and on average 84\% of people prefer our reconstruction. Some example images are shown in ~\cref{fig:coco-example} and \cref{fig:comp_ntu}. See supplementary for details on the user study and more qualitative examples.

\textbf{Comparison with Weng \etal \cite{weng2020holistic}.} One limitation of \cite{weng2020holistic} is that they use some indoor scene layout assumptions, which makes it unsuitable to compare on 'in the wild' COCO images. Instead, we only test it on the NTU-RGBD dataset. Similar to the comparison with PHOSA, we released a user study on AMT and ask 50 users to select which reconstruction is better. Results show that our method is preferred by 94\% people. Our method clearly outperforms \cite{weng2020holistic}, see two examples in \cref{fig:comp_ntu} and more examples in supplementary.

\subsection{Comparison with a human reconstruction method}
Our method takes SMPL pose predicted from FrankMocap\cite{rong2020frankmocap} as initialisation and fine tunes it together with object using network predictions. We compare the human reconstruction performance with FrankMocap and PHOSA\cite{zhang2020phosa} in \cref{table:comp-mocap}. It can be seen that our joint reasoning of human and object does not deteriorate the human reconstruction. On the other hand,  it slightly improves the human reconstruction as it takes the interacting object into account. 
\begin{table}
    \centering
    \begin{tabular}{l c c c}
    \toprule
     Metric (cm) & PHOSA & FrankMocap & Ours \\
    \midrule
    v2v $\downarrow$ & 6.90 $\pm$ 2.68 & 6.91 $\pm$ 2.75 & {\bf 5.46 $\pm$ 3.47}\\
    Chamfer $\downarrow$ & 6.11 $\pm$ 1.98 & 6.12 $\pm$ 2.00 & {\bf 5.27 $\pm$ 1.98}\\
    \bottomrule
    \end{tabular}
    \caption{Mean vertex to vertex (v2v) and Chamfer distance for human reconstruction results after Proscrustes alignment on SMPL mesh. Our method that jointly reasons about human and object performs on par with human only FrankMocap~\cite {rong2020frankmocap}.}
    \label{table:comp-mocap}
\end{table}

\subsection{Pixel-aligned learning on real data}\label{sec-smpl-centering}
CHORE is trained with the depth-aware scaling strategy described in \cref{sec:centering} for effective pixel aligned implicit surface learning on real data. 
To evaluate its importance, we compare with two baseline models trained with weak perspective and direct perspective (no depth-aware scaling on GT data) camera. Qualitative comparison of the neural reconstruction is shown in \cref{fig:orth_vs_persp}. We further fit SMPL and object meshes using these baseline models (with object pose prediction) on \behave{} test set  and compute the errors for reconstructed meshes, see \cref{table:ablate-study}b) and c). 
It can be clearly seen that our proposed training strategy allows effective learning of implicit functions and significantly improves the reconstruction. 

\subsection{Contacts modeling} \label{sec:contact}

We have argued that contacts between human and the object are crucial for accurately modelling interaction. Quantitatively, the contact term $E^c_{data}$ only reduces the object error slightly (\cref{table:ablate-study}f) and g) as contacts usually constitute a small region of the full body. Nevertheless, it makes significant difference qualitatively, as can be seen in Fig. 7. Without the contact information, the object cannot be fitted correctly to the location where contact is happening. Our contact prediction helps snap the object to the correct contact location, leading to a more physically plausible and accurate reconstruction. 
\noindent %
\begin{minipage}[]{\textwidth}
  \begin{minipage}[b]{0.49\textwidth}
    \centering
    \includegraphics[width=1.0\linewidth]{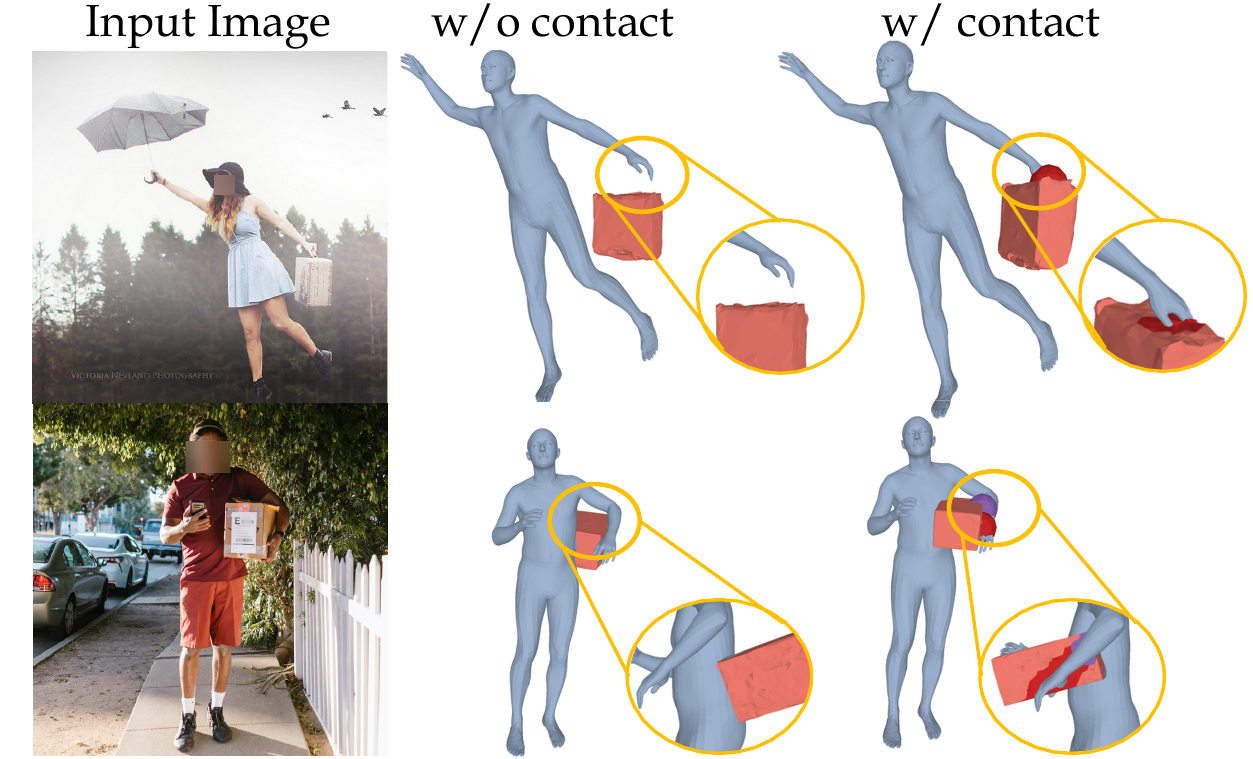}
    \label{fig:contacts}
    \captionof{figure}{Accurate human and object reconstruction requires modeling contacts explicitly. Without contact information, the human and the objects are not accurately aligned and the resulting reconstruction is not physically plausible.}
  \end{minipage}
  \hfill
  \begin{minipage}[b]{0.49\textwidth}
    \centering
    \begin{tabular}{l c c c}
    \toprule
    Methods  & SMPL $\downarrow$ & Object $\downarrow$\\
    \midrule
    a. neural recon. &  14.06   & 15.34\\
    b. weak persp. &  9.14  & 25.24 \\
    c. direct persp. &  7.63  & 15.58 \\
    d. w/o $f^R$ &  6.58  & 17.54 \\
    e. w/o $f^c$ &  6.26  & 14.25 \\
    f. w/o contact &  {\bf 5.51 }  &  10.86 \\
    g. { full model} & 5.58  & {\bf 10.66 }\\
    \bottomrule
    \end{tabular}
      \captionof{table}{Ablation studies. 
      We can see that neither weak perspective (b) nor direct perspective camera (c) can work well on real data. Our proposed training strategy, together with rotation and translation prediction achieves the best performance.}
      \label{table:ablate-study}
    \end{minipage}
    \end{minipage}

\subsection{Other ablation studies} \label{sec:ablate-object-center}
We further ablate our design choices in \cref{table:ablate-study}. Note that all numbers in \cref{table:ablate-study} are Chamfer distances (cm) and computed on meshes after joint fitting except a) which is computed on the neural reconstructed point clouds. 
We predict object rotation and translation to initialize and regularize object fitting, which are important for accurate reconstruction. In \cref{table:ablate-study}, it can be seen that fitting results without rotation d) or without translation initialization e) leads to worse perofrmance compared to our full model g). Without initialization the object gets stuck in local minima. See supplementary for qualitative examples. 

We also compute the Chamfer distance of the neural reconstructed point clouds (\cref{eq:udf_recon}) using alignments computed from fitted meshes. It can be seen that this result is much worse than the results after joint fitting (\cref{table:ablate-study}g). This is because network predictions can be noisy in some regions while our joint fitting incorporate priors of human and object making the model more robust.

\section{Limitations and Future Work}
\label{sec:limitations}
We introduced the first learning based method  to reconstruct 3D human and object from images. Although the results are promising, there are some limitations. 
For instance, we assume known object template for the objects shown in the images. Future works can remove this assumption by selecting template from 3D CAD database or reconstructing shape directly from image. Non-rigid deformation, multiple objects, multi-person interactions are exciting avenues for future research. Please also refer to supplementary for discussions of failure cases.

\section{Conclusion}
In this work we presented CHORE, the first method to \emph{learn} a \emph{joint} reconstruction of human, object and contacts from single images. CHORE combines powerful neural implicit predictions for human, object and its pose with robust model-based fitting. 
Unlike prior works on pixel-aligned implicit surface learning, we learn from real data instead of synthetic scans, which requires a different training strategy. We proposed a simple and elegant depth-aware scaling that addresses the depth-scale ambiguity while preserving pixel alignment.

Experiments on three different datasets demonstrate the superiority of CHORE compared to the SOTA.
Quantitative experiments on \behave{} dataset evidence that joint reasoning leads to a remarkable improvement of over $50\%$ in terms of Chamfer distance compared to PHOSA. 
User studies on the NTU-RGBD and the challenging COCO dataset (no ground truth) show that our method is preferred over PHOSA by 84\% and 72\% users respectively.
We also conducted extensive ablations, which reveal the effectiveness of the different components of CHORE, and the proposed depth-aware scaling for pixel-aligned learning on real data. Our code and models are released to promote further research in this emerging field of single image 3D human-object interaction capture.

{\subsubsection{Acknowledgements.} \small We would like to thank RVH group members \cite{rvh_grp} for their helpful discussions. Special thanks to Beiyang Li for supplementary preparation. This work is funded by the Deutsche Forschungsgemeinschaft (DFG, German Research Foundation) - 409792180 (Emmy Noether Programme,
project: Real Virtual Humans), and German Federal Ministry of Education and Research (BMBF): Tübingen AI Center, FKZ: 01IS18039A. Gerard Pons-Moll is a Professor at the University of Tübingen endowed by the Carl Zeiss Foundation, at the Department of Computer Science and a member of the Machine Learning Cluster of Excellence, EXC number 2064/1 – Project number 390727645.
}

\bibliographystyle{splncs04}
\bibliography{egbib}
\end{document}